\pdfoutput=1

\documentclass[11pt]{article}

\usepackage[]{EACL2023}

\usepackage{times}
\usepackage{latexsym}

\usepackage[T1]{fontenc}

\usepackage[utf8]{inputenc}

\usepackage{microtype}

\usepackage{inconsolata}

\usepackage{booktabs}
\usepackage{multirow}
\usepackage[pdftex]{graphicx}
\usepackage{xcolor}
\usepackage{color}
\usepackage[normalem]{ulem}
\usepackage{url}
\usepackage[utf8]{inputenc}

\usepackage{amsmath}
\usepackage[english]{babel}
\usepackage{hyperref}
\addto\captionsenglish{%
}
\addto\extrasenglish{%
}

\newcommand{\blue}[1]{\colorbox[rgb]{0.92,0.96,1}{#1}} 
\newcommand{\red}[1]{\colorbox[rgb]{1,0.9,0.9}{#1}} 
\newcommand{\prompt}{\texttt}

\title{Evaluating the Robustness of Discrete Prompts}

\author{
    Yoichi Ishibashi$^1$ \quad
    Danushka Bollegala$^{2}$ \quad
    Katsuhito Sudoh$^1$ \quad
    Satoshi Nakamura$^1$ 
    \\
    $^1$ Nara Institute of Science and Technology \quad
    $^2$ University of Liverpool
    \\
    \texttt{\{ishibashi.yoichi.ir3, sudoh, s-nakamura\}@is.naist.jp} \\
    \texttt{danushka@liverpool.ac.uk}
  }

\begin{document}
\maketitle
\begin{abstract}
Discrete prompts have been used for fine-tuning Pre-trained Language Models for diverse NLP tasks.
In particular, automatic methods that generate discrete prompts from a small set of training instances have reported superior performance.
However, a closer look at the learnt prompts reveals that they contain noisy and counter-intuitive lexical constructs that would not be encountered in manually-written prompts.
This raises an important yet understudied question regarding the \emph{robustness} of automatically learnt discrete prompts when used in downstream tasks.
To address this question, we conduct a systematic study of the robustness of discrete prompts by applying carefully designed perturbations into an application using AutoPrompt and then measure their performance in two Natural Language Inference (NLI) datasets.
Our experimental results show that although the discrete prompt-based method remains relatively robust against perturbations to NLI inputs, they are highly sensitive to other types of perturbations such as shuffling and deletion of prompt tokens. 
Moreover, they generalize poorly across different NLI datasets.
We hope our findings will inspire future work on robust discrete prompt learning.\footnote{Our codes and the adversarial NLI dataset are available at \url{https://github.com/LivNLP/prompt-robustness}}
\end{abstract}

\section{Introduction}
\label{sec:intro}
Pre-trained Language Models (PLMs) have been successfully adapted to a wide range of Natural Language Processing (NLP) tasks using \emph{prompt-based} learning~\cite{Radford:2018,GPT-2,GPT3,petroni-etal-2019-language} such as sentiment classification~\cite{gao-etal-2021-making}, natural language inference (NLI)~\cite{schick-schutze-2021-exploiting,schick-schutze-2022-true}, relation extraction~\cite{shin-etal-2020-autoprompt}, cross-lingual inference~\cite{qi-etal-2022-enhancing}.
However, manually writing prompts that generalize well is very challenging for several reasons such as (a) it might not always be possible to recruit domain-expert human annotators, 
(b) human annotators might not be able to cover all corner cases by writing prompts, and
(c) there can be disagreements between human annotators regarding the coverage of a particular prompt.
To address these challenges, automatic learning of discrete prompts has been proposed such as AdvTrigger~\cite{wallace-etal-2019-universal}, AutoPrompt~\cite[\textbf{AP};][]{shin-etal-2020-autoprompt}, WARP~\cite{hambardzumyan-etal-2021-warp}, and RLPrompt~\cite{DBLP:journals/corr/abs-2205-12548}.

\begin{table*}[t]
    \centering
    \small
    \begin{tabular}{lllc}
        \toprule
         \textbf{Relation} &  \textbf{Method} & \textbf{Prompt} & \textbf{P@1} \\ \midrule
         \textsf{native-language-of} (P103)   & Manual            & \prompt{The native language of [X] is [Y]} & 74.54\\
                                            & AP BERT   & \prompt{[X]PA communerug speaks proper [Y]} & \textbf{84.87}\\
                                            & AP RoBERTa & \prompt{[X]neau optionally fluent!?\" traditional [Y]} & 81.61\\ \midrule
         \textsf{profession-of} (P106) & Manual & \prompt{[X] is a [Y] by profession} & 0.73 \\
                                            & AP BERT & \prompt{[X] supporters studied politicians musician turned [Y]} & 15.83 \\
                                            & AP RoBERTa & \prompt{[X] (), astronomers businessman·former [Y]} & \textbf{19.24} \\ \midrule 
         \textsf{music-played-by} (P136) & Manual & \prompt{[X] plays [Y] music} & 0.7\\
                                         & AP BERT & \prompt{[X] freaking genre orchestra fiction acid [Y]} & \textbf{59.95} \\
                                          & AP RoBERTa & \prompt{[X] blends postwar hostage drama sax [Y]} & 52.97 \\
         \bottomrule
    \end{tabular}
    \caption{Examples of prompts learnt by AP for the fact retrieval task for BERT and RoBERTa PLMs and the human-written manual prompts. T-REx relation ids are shown with brackets for each relation type. Precision@1 (P@1) scores are shown when each prompt is used in fact retrieval.}
    \label{tbl:autoprompt-examples}
\end{table*}

Although discrete prompt learning methods have achieved good performance in numerous downstream tasks by automatically learnt prompts, such automatic prompts seem to be significantly different from the manually-written ones.
For example, \autoref{tbl:autoprompt-examples} shows manually-written and AP-learnt prompts for fact retrieval~\cite{petroni-etal-2019-language}.
We see that the AP-learnt prompts for BERT~\cite{devlin-etal-2019-bert} and RoBERTa~\cite{RoBERTa} outperform the manual prompts in precision\@1 (P@1) scores.
However, the AP-learnt prompts contain various counter-intuitive language constructs such as punctuation (e.g. `(', `?', `!', `)'), spelling errors (e.g. \emph{commuenrug}) etc., which seem unrelated to the target relation.
Similar cases can be observed for AP-learnt prompts for other tasks as well (see Appendix in \newcite{shin-etal-2020-autoprompt}).
It is unrealistic that a human annotator would be able to write such prompts even if they were able to see the same training instances as used by automatic methods.

Considering the fact that discrete prompt learning methods are trained in a few-shot setting where they use only a small number of training instances, the seemingly counter-intuitive nature of the discrete prompts learnt by automatic methods raises concerns about their robustness.
For example, \emph{How will the performance of a target task change if we add small random perturbations to the prompts learnt by AP?} and \emph{Whether the prompts learnt by AP generalize to out-of-domain data?}.
To study these issues, in this paper we evaluate the robustness of discrete prompts learnt by automatic prompt learning methods and compare that with manually-written prompts and direct fine-tuning of PLMs.

An evaluation of the robustness of discrete prompts is important for two main reasons.
First, given that discrete prompt learning methods are learning those prompts from a small set of training instances, it is important that they cover the core patterns that generalize to the target task and not simply capture some random artefacts in the training samples.
Second, unlike embedding-based continuous prompts~\cite{li-liang-2021-prefix,lester-etal-2021-power}, discrete prompts~\cite{wallace-etal-2019-universal,shin-etal-2020-autoprompt,DBLP:journals/corr/abs-2205-12548} are represented in natural language and supposed to be interpretable.
However, if a discrete prompt learning method is less robust, a seemingly harmless perturbation such as removing a punctuation character can significantly alter the performance of a downstream task.

In contrast to the numerous work that has used prompts for fine-tuning PLMs, to the best of our knowledge, the robustness of discrete prompts to random or adversarial perturbations has not been systematically studied.
To address this gap, we use AP as a concrete example of a widely-used method and evaluate its robustness under different types of carefully designed perturbations.
However, we note that our perturbation techniques are not limited to AP and can be used for any discrete prompt learning method.
We compare the performance of AP-learnt prompts against fine-tuning using Manually-written Prompts (MP), and Head-based Fine-Tuning (HFT), where we fine-tune both the classifier head and the PLM parameters. 

From our evaluation, we find several interesting facts about the robustness of discrete prompts as summarized below.
\begin{itemize}
    \item Overall, when the number of training instances is increased, MP outperforms both AP and HFT on CB~\cite{de23commitmentbank} and MNLI~\cite{DBLP:conf/naacl/WilliamsNB18}, two independent benchmark datasets for NLI (\autoref{sec:exp:datasize}).
    In particular, the performance of AP on MNLI is much worse than that on CB. This is in contrast to the superior performance of AP on SICK-E~\cite{DBLP:conf/lrec/MarelliMBBBZ14}, another NLI dataset, as reported by \newcite{shin-etal-2020-autoprompt}.
    
    \item Moreover, we see a performance drop when we use discrete prompts learnt from CB for MNLI and vice-versa (\autoref{sec:exp-cross-data}).
    These results indicate that the performance of discrete prompts learnt by AP is highly dataset-dependent and such discrete prompts do not generalize well across datasets.
    
    \item Compared to MP, AP-learnt discrete prompts turn out to be highly sensitive to the ordering of prompt tokens (\autoref{sec:exp:reorder}).
    
    \item Random deletion of prompt tokens decreases performance in both AP and MP (\autoref{sec:exp:deletion}).
    
    \item  We create an adversarial NLI dataset from randomly-sampled test instances from MNLI and CB, and manually modify the hypothesis sentences with keeping the corresponding premise sentences unchanged, such that (a) the target label would not change, and (b) would reverse an entailment label to a contradiction (or vice-versa).
    Both AP and MP remain relatively robust against the perturbations that do not change the target label, but the performance of MP drops significantly in the label-changing setting (\autoref{sec:exp:input-perturbation}).
    This shows that AP is relatively more robust against adversarial perturbations than MP, which explains AP's superior performance in various tasks. 
\end{itemize}

\section{Related Work}
\paragraph{Prompting Methods:}
Prompting or \emph{in-context-learning} has received wide attention as an efficient method to extract knowledge from PLMs ~\cite{GPT3,petroni-etal-2019-language,cui-etal-2021-template}.
However, to manually write prompts one must possess task-specific domain knowledge.
As an alternative, methods that can automatically learn prompts from training data have been proposed.
Two distinct types of prompts have been learnt in prior work:
discrete prompts (learns lexical sequences), and continuous prompts (learns embeddings).
Continuous prompts~\cite{li-liang-2021-prefix,lester-etal-2021-power} are parameter efficient because they learn generalizable task-specific embeddings, with performance comparable to PLM fine-tuning.
However, continuous prompts cannot be learnt when a PLM is publicly unavailable and the only access to it is via an API~\cite{GPT3}.
Moreover, compared to discrete prompts, continuous prompts are difficult to interpret.
Learning discrete prompts~\cite{wallace-etal-2019-universal,shin-etal-2020-autoprompt,DBLP:journals/corr/abs-2205-12548} does not suffer from these limitations of continuous prompts and can be used with diverse NLP tasks.
Especially, fine-tuning massive PLMs has become computationally costly, which has made discrete prompt learning an attractive alternative.

\paragraph{Analysis of Prompting Methods:}
Prior work has analyzed prompts from various viewpoints. 
\citet{DBLP:conf/naacl/ScaoR21} studied the effect of training dataset size on fixed-prompt PLM fine-tuning and head-based fine-tuning and found that prompting is often worth 100s of instances on average across classification tasks.
\citet{kavumba-etal-2022-prompt} showed that the performance of prompt-based models varies significantly depending on the surface cues in the sentence.
\citet{lu-etal-2022-fantastically} found that ordering of task input significantly affects the performance.
\citet{utama-etal-2021-avoiding} focused on the reliance on lexical overlap in sentence pair classification and showed that prompt-based models fail to make predictions dependent on the lexical overlap.
To the best of our knowledge, the robustness of discrete prompts under different types of perturbations has not been studied in prior work, which is the main focus of this paper.

\section{Experiments}
\label{sec:experiments}
Let us first describe experimental settings common to all experiments.

\paragraph{Prompting and Fine-Tuning Methods: }
We compared the following methods. 

\begin{itemize}
    \item \textbf{AutoPrompt}~\cite[\textbf{AP};][]{shin-etal-2020-autoprompt} is a representative method of discrete prompt learning.
    The learning strategy is based on fill-in-the-blank task~\cite{devlin-etal-2019-bert}.
    First, a manually created prompt template (e.g., \prompt{[X] <MASK> <T> ... <T> [Y]}) is given, and a prompt token (called a trigger token) is learnt by replacing \prompt{<T>}, which is a special token representing a trigger token.
    In the search for trigger tokens, the probability of \prompt{<MASK>} is converted into class probability by using label tokens (e.g., \{`\emph{nobody}', `\emph{nor}'\} for contradiction~\cite{shin-etal-2020-autoprompt}), and trigger tokens are searched by gradient-guided search~\cite{wallace-etal-2019-universal} to find a candidate set consisting of trigger tokens from a vocabulary of the language model. 
    As a template for NLI, we used the one given by \citet{shin-etal-2020-autoprompt}, and the prompt tokens were learnt from the training dataset. 
    In our experiments, we used the official implementation.\footnote{\url{https://github.com/ucinlp/autoprompt}}
    
    \item \textbf{Manually-written Prompts}~\cite[\textbf{MP};][]{schick-schutze-2021-exploiting} is a method for fine-tuning the entire masked language model with training data using manually-written prompts as the input and predicting the \prompt{<MASK>} tokens for the labels (e.g., `\emph{yes}' for entailment).
    We used the template \prompt{\{hypothesis\}? | <MASK>, \{premise\}} and  verbalizer (`\emph{yes}' for entailment, `\emph{no}' for contradiction, `\emph{maybe}' for neutral) following prior work \cite{schick-schutze-2021-exploiting,DBLP:conf/naacl/ScaoR21}.
    \citet{schick-schutze-2021-exploiting} proposed an ensemble-based method with multiple rounds of fine-tuning using different templates.
    However, because a single template is used in AP, for a fair comparison in our experiments, we fine-tuned a PLM using one MP template.

    \item \textbf{Head-based Fine-Tuning}~\cite[\textbf{HFT};][]{devlin-etal-2019-bert} fine-tunes the PLM with a classifier head.
    We report the head-based results trained by \newcite{DBLP:conf/naacl/ScaoR21}. 
    They trained HFT with a low learning rate ($10^{-5}$) and always with a large number of steps (at least 250), following the recommendations in prior work~\cite{DBLP:conf/iclr/MosbachAK21,DBLP:conf/iclr/0007WKWA21}.
    Note that HFT is not a prompt-based method, so it was excluded from some experiments on the robustness of discrete prompts.
\end{itemize}

\paragraph{Datasets:}
We used NLI as an evaluation task to compare the robustness of discrete prompting methods.
The NLI task has been used in multiple previous studies to evaluate and/or propose novel prompt learning methods because it is a fundamental task related to many NLP applications~\cite{shin-etal-2020-autoprompt,DBLP:conf/naacl/ScaoR21,DBLP:conf/naacl/WebsonP22}.
It is important to use the same NLI task and datasets in our experiments to facilitate fair comparisons and reach reproducible conclusions. 
We used the two datasets: CommitmentBank \cite[\textbf{CB};][]{de23commitmentbank}\footnote{\url{https://super.gluebenchmark.com/tasks}} (a corpus of short texts), and Multi-Genre Natural Language Inference Corpus \cite[\textbf{MNLI};][]{DBLP:conf/naacl/WilliamsNB18}\footnote{\url{https://cims.nyu.edu/~sbowman/multinli/}} (a crowdsourced collection of sentence pairs for NLI).
Each sentence pair is labelled with \emph{entailment}, \emph{neutral}, or \emph{contradiction}.

\paragraph{PLM:}
In our experiments, we used the same pre-trained language model to evaluate AP, MP, and HFT equally.
Specifically, we used RoBERTa-large (355M parameters) \footnote{\url{https://huggingface.co/roberta-large}}~\cite{RoBERTa}, which has been used in much prior work in prompt learning~\cite{shin-etal-2020-autoprompt,DBLP:conf/naacl/ScaoR21}.
The PLM was trained on five datasets, including BookCorpus\footnote{\url{https://yknzhu.wixsite.com/mbweb}}, English Wikipedia\footnote{\url{https://en.wikipedia.org/wiki/English_Wikipedia}}, CC-News\footnote{\url{https://commoncrawl.org/2016/10/news-dataset-available/}}, OpenWebText\footnote{\url{https://github.com/jcpeterson/openwebtext}}, and Stories\footnote{\url{https://arxiv.org/abs/1806.02847}}.
The texts were tokenised using a byte-level Byte-Pair Encoding \cite[BPE;][]{DBLP:conf/acl/SennrichHB16a} vocabulary of size 50,000.

\paragraph{Evaluating the Robustness of Prompts: }
We used \emph{rate of degradation} (\textbf{RoD})~\cite{Meyers:2020} to evaluate robustness, which is defined as the decrease in accuracy of the target task due to the perturbations added to the prompt.
If the RoD of a model is small after the inclusion of a perturbation, the model is considered to be robust against that perturbation.
Specifically, we first calculate the respective accuracies $\textrm{acc}_x$ and $\textrm{acc}_{x^\ast}$ on the same evaluation set for both prompt $x$ and its perturbated version $x^{\ast}$.
Using the average accuracies $\textrm{avg-acc}_x$ and $\textrm{avg-acc}_{x^\ast}$ over $M$ prompts ${x_1, ..., x_M}$, we calculate the RoD as $(\textrm{avg-acc}_x - \textrm{avg-acc}_{x^\ast}) / \textrm{avg-acc}_x = 1 - \textrm{avg-acc}_{x^\ast} / \textrm{avg-acc}_x$.

\subsection{Effect of the Training Dataset Size}
\label{sec:exp:datasize} 
Before moving on to robustness experiments, we first investigate the number of training instances on which AP and MP perform best, and used the best-performing AP and MP to evaluate their robustness in the subsequent experiments.

\paragraph{Experimental Settings:} 
We gradually increased the size of the training dataset following the experimental setup of~\citet{DBLP:conf/naacl/ScaoR21}. 
Specifically, we experimented with randomly sampled subsets of the training dataset having varying numbers of instances in $\{10, 15, 20, 30, 50, 70, 100, 150, 200\}$.
Because the performance of few-shot learning methods often varies due to the high feature variance in the training data, we randomly sampled four subsets per each dataset size and used them independently for training the models\footnote{NVIDIA RTX A5000 was mainly used.} (i.e. trigger tokens and label tokens for AP, or fine-tuned language model for MP and HFT) for each subset and report the average accuracy on the validation data for the four models ($M = 4$).
We used the matched (example from the same source as the training set) validation set for MNLI.
For CB, we held out 50 training instances for development as in \citet{DBLP:conf/naacl/ScaoR21} and evaluated the original validation set as test data.

We searched for the optimal values for the following hyperparameters: the number of trigger tokens in \{3, 5, 10\}, the number of label tokens in \{3, 5, 10\}, and the number of tokens in a candidate set in \{10, 50\}. 
We evaluated the test accuracy using the hyperparameters that had the highest accuracy on the validation data for each dataset size.
In the training of MP, we used AdamW optimizer~\cite{DBLP:conf/iclr/LoshchilovH19} with an initial learning rate of $10^{-5}$ and a learning step of 1,000 following \citet{DBLP:conf/iclr/MosbachAK21}.

\begin{table*}[t]
    \centering
    \small
    \begin{tabular}{ccccccc}
        \toprule
        \textbf{Method} & \textbf{\#Train} & \textbf{Template} & \textbf{\#Prompt tokens} & \textbf{\#Label tokens per class} & \multicolumn{2}{c}{\textbf{Avg. accuracy}} \\
        & & & & & \textbf{CB} & \textbf{MNLI} \\
        \midrule
        AP & 200 & \prompt{\red{p} <MASK> \blue{<T> ... <T>} \red{h}} & 10 & 3 & 68.3 & 37.7 \\ 
        MP & 200 & \prompt{\red{h}\blue{? |} <MASK>\blue{,} \red{p}} & 3 & 1 & \uline{95.1} & \uline{65.5} \\ 
        HFT & - & \prompt{<CLS> \red{p} <SEP> \red{h}} & 0 & - & - & - \\ 
        \bottomrule
    \end{tabular}
    \caption{
    The average accuracy of the experiment with four training subsets of 200 instances.
    \red{Red} represents the task inputs, \red{\prompt{h}} represents the hypothesis, \red{\prompt{p}} represents the premise, \blue{blue} represents the prompt tokens, and \blue{\prompt{<T>}} represents a trigger token.
    Unreported values were marked with `-'. 
    }
    \label{tab:pre-train-best}
\end{table*}

\begin{figure}[t]
  \centering
    \includegraphics[clip, width=8cm]{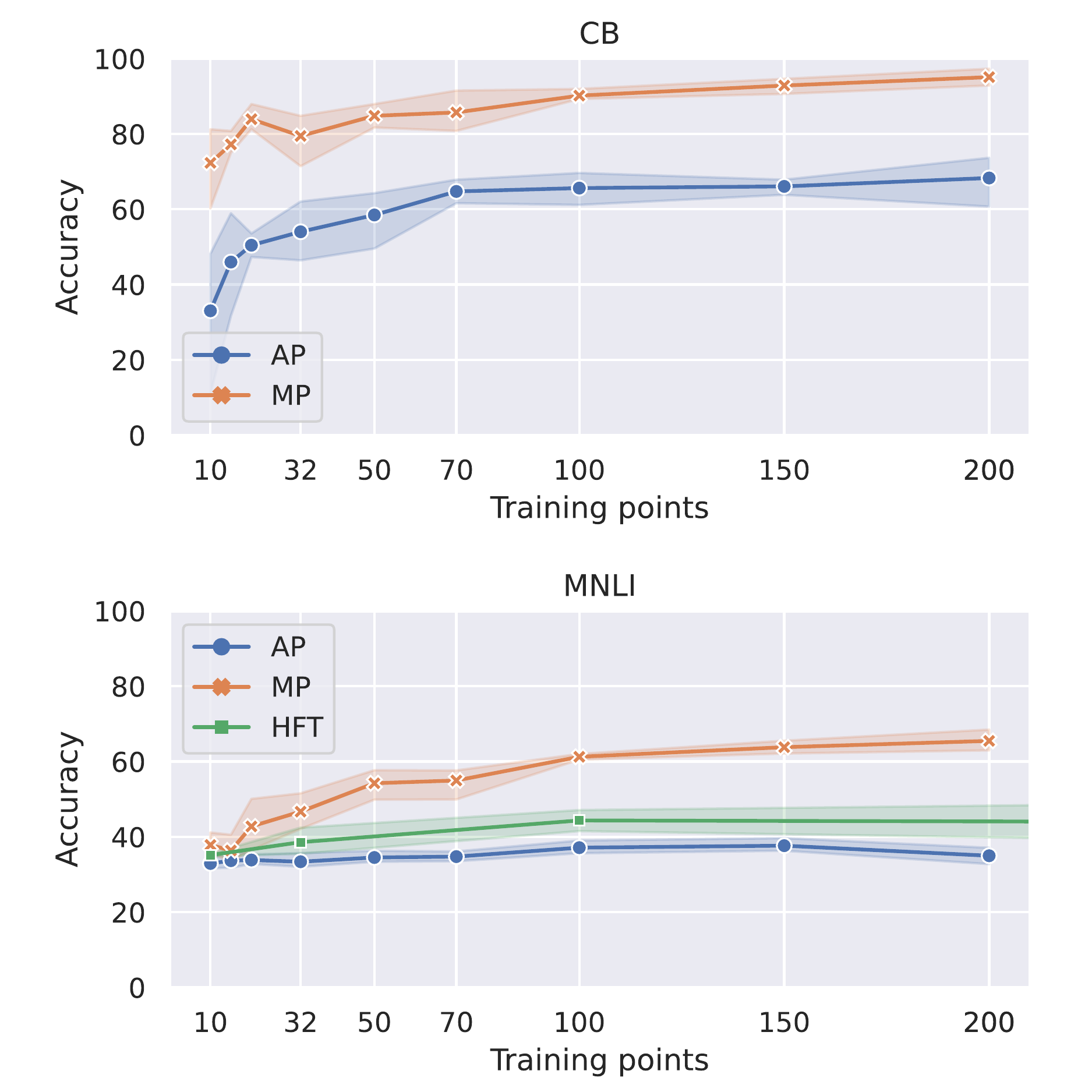}
    \caption{
    Performance of AutoPrompt (AP), Manually-written Prompt (MP), and Head-based Fine-Tuning (HFT) on the scale of dataset size for CB and MNLI.
    Means and their 95\% confidence intervals are plotted.
    The accuracy of HFT for dataset size for CB was not plotted because the accuracy was not reported.
    }
    \label{fig:pre-train}
\end{figure}

\paragraph{Main Results:}
\autoref{fig:pre-train} shows the performance\footnote{HFT results were obtained from \citet{DBLP:conf/naacl/ScaoR21}, F1-macro for CB and accuracy for MNLI.} against the training dataset size.
We see that in both CB and MNLI \textbf{MP is always superior to AP}.
For example, with a dataset of size 200, AP and MP achieved the best accuracy in CB, MP's accuracy was 92.7\%, while that of AP was lower at 54.2\%.

Our results also suggest that \textbf{the performance of discrete prompts learnt by AP is highly dataset dependent.}
\citet{shin-etal-2020-autoprompt} reported results for AP and HFT on SICK-E~\cite{DBLP:conf/lrec/MarelliMBBBZ14}, which is another NLI dataset.
They concluded that AP was always superior to HFT up to training dataset sizes of 1,000 for the same RoBERTa-large PLM that we use.
However, our experiments show the opposite trend (i.e. HFT is superior to AP).
This suggests that even if AP is superior to HFT on a given dataset, it is not guaranteed to be superior in a different dataset for the same task.
This may be due to the differences in the domain and annotation guidelines for each dataset.
For example, the accuracy of MNLI was quite low on AP, which contrasts with that of CB.
This result suggests that the discrepancies in domains and annotation guidelines make it difficult for AP to perform consistently.

\paragraph{Best Prompts:}
\autoref{tab:pre-train-best} shows the average accuracy of models trained on 200 instances that performed well in both CB and MNLI.
Note that there are four training subsets for each dataset size, resulting in corresponding four trained AP prompts and four PLMs fine-tuned by MP. \footnote{We show the four best prompts learnt by AP in \autoref{sec:supplementary}.}
In the robustness evaluations in \autoref{sec:exp:reorder} through \autoref{sec:exp:input-perturbation}, we used these learnt APs and MPs.
In this paper, (a) trigger tokens learnt by AP, and (b) manually-written prompts excluding the task inputs and  mask tokens are collectively referred to as the \emph{prompt tokens}.

\subsection{Token Reordering}
\label{sec:exp:reorder}
As seen from \autoref{tbl:autoprompt-examples}, compared to MPs where the ordering of tokens in a prompt is manually determined, discrete prompts learnt by AP appear to have no obvious ordering among their tokens.
To empirically investigate the importance of the token order in a discrete prompt, we conduct an experiment where we randomly shuffle the prompt tokens and measure the effect on the downstream task performance.

\paragraph{Experimental Procedure:}
Given a discrete prompt, we first randomly reordered its prompt tokens (e.g. shaded in blue in \autoref{tab:pre-train-best}).
Next, we used the reordered prompt with the PLM to make entailment predictions for the test instances in the CB and MNLI datasets.
Finally, the entailment prediction accuracy (Acc) obtained with the reordered prompts was computed.
We repeated this evaluation 10 times for each prompt and report the averaged values and the corresponding RoD values.

\paragraph{Main Results:}
From \autoref{tab:reorder_result} we see that the accuracy drops for both AP and MP when the prompt tokens are randomly reordered.
In particular, the accuracy of AP drops significantly compared to that of MP.
For example, the accuracy of AP on CB drops by ca. 14\% due to token reordering, while that for MP drops only by ca. 2\%.
Intuitively, one would expect that changing the order of prompt tokens in MP would result in a significant drop in accuracy because the meaning of the prompts would change.
However, we see that this is not the case. 
This result shows that \textbf{the discrete prompts learnt by AP strongly rely on the token order}.

\begin{figure}[t]
  \centering
    \includegraphics[clip, width=7.5cm]{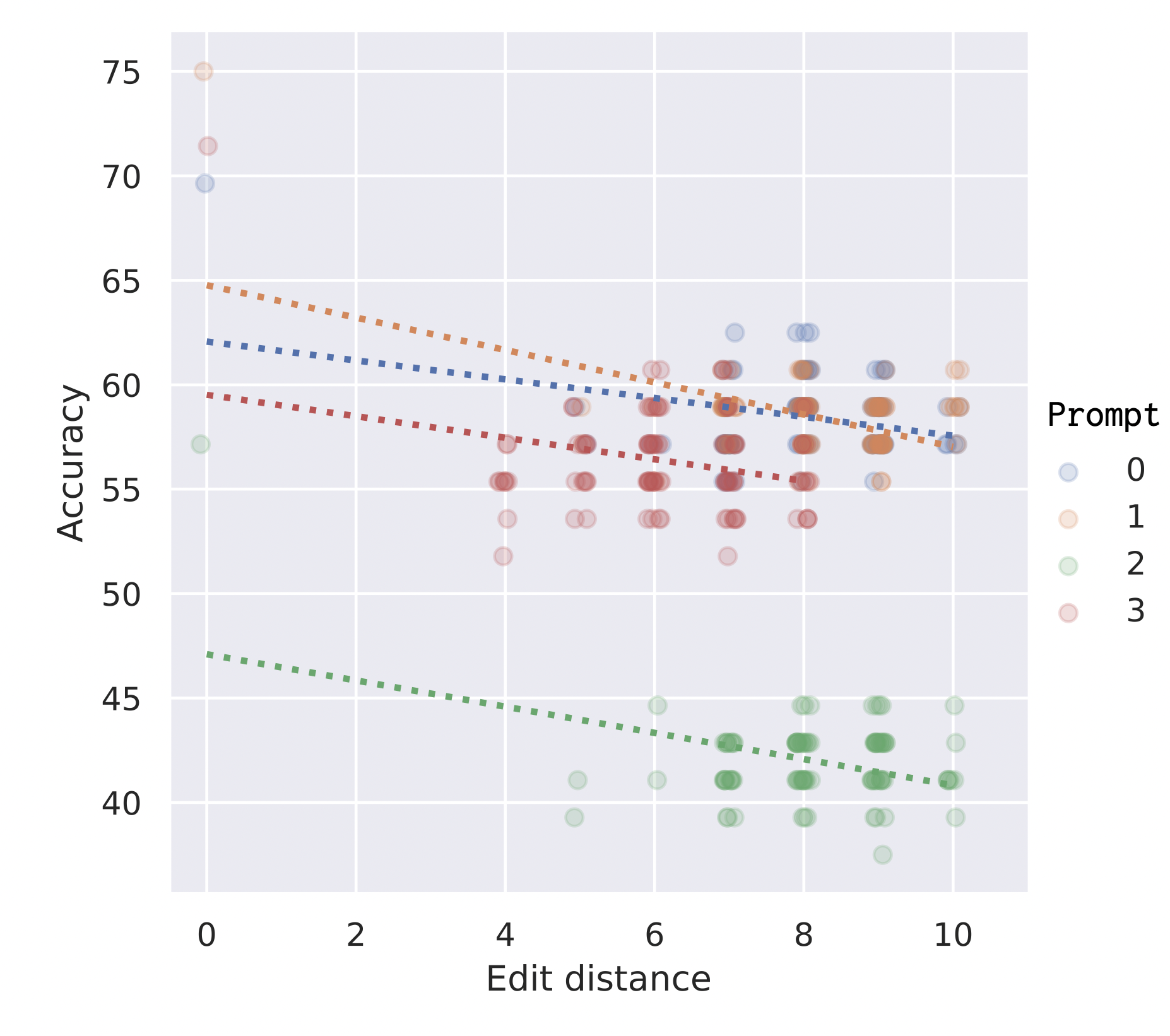}
    \caption{Edit distance and accuracy of the reordered trigger tokens. We evaluated them on the validation data of CB. 
    The prompts numbered 0 through 3 each represent the four prompts learnt by AP (shown in \autoref{tab:best-ap-prompts-cb}).
    Note that a point with an edit distance of zero indicates accuracy with the original trigger token.
    }
    \label{fig:scatter_reodering}
    \vspace{-3mm}
\end{figure}

\paragraph{Additional Analysis:}
To further analyze the relationship between the level of perturbation introduced by reordering prompt tokens in AP and its effect on the performance, we computed the token-level edit distance \citep[Levenshtein distance;][]{levenshtein1966binary} between each prompt and its token-shuffled version as shown in \autoref{fig:scatter_reodering}.
For all four AP prompts, we see that the accuracy drops when the perturbation noise (i.e. measured by edit distance) increases.
This reconfirms the lack of robustness in discrete prompts learnt by AP to the random shuffling of prompt tokens.

\begin{table}[t]
    \centering
    \small
    \begin{tabular}{clllccc}
        \toprule
        \textbf{Method} & \textbf{Metrics} & \textbf{CB} & \textbf{MNLI} \\
        \midrule
        \multirow{2}{*}{AP} & Acc & \textbf{54.2} & \textbf{34.3} \\ 
                            & RoD & \textbf{0.21} & \textbf{0.10} \\ 
        \midrule
        \multirow{2}{*}{MP} & Acc & \uline{92.7} & \uline{59.3} \\
                            & RoD & \uline{0.03} & \uline{0.09} \\ 
        \bottomrule
    \end{tabular}
    \caption{
    Performance of reordered prompts.
    Acc denotes accuracy; RoD denotes the RoD from before the reordering (\autoref{tab:pre-train-best}).
    The largest drops in accuracy are \textbf{bolded} and the smallest drops are \uline{underlined} for each method and dataset.
    AP relies more strongly on word order than MP.
    }
    \label{tab:reorder_result}
\end{table}

\subsection{Token Deletion} 
\label{sec:exp:deletion}
As seen from \autoref{tbl:autoprompt-examples}, the discrete prompts learnt by AP perform better than MP.
However, it is often difficult to determine the importance of prompt tokens to the target task due to their lack of interpretability (e.g. prompt token `\emph{neau}' in \autoref{tbl:autoprompt-examples}).
To understand the significance of individual prompt tokens to the overall discrete prompt, we conducted an experiment where we systematically deleted one or more prompt tokens at various positions from a given discrete prompt and measure the drop (if any) in the performance of the NLI task.

\paragraph{Experimental Procedure:}
We evaluated two settings of prompt deletion: \emph{single} and \emph{multiple} token deletion.
In the single token deletion setting, we deleted one token at different positions in a given prompt.
For AP, we repeated this with each of the four discrete prompts (shown in \autoref{tab:pre-train-best}) and report the average accuracy.
In the multiple token deletion setting, we delete $n \in \{1, 3, 5, 7\}$ prompt tokens following three strategies:
\emph{Random-deletion} deletes $n$ prompt tokens randomly,
\emph{Front-deletion} deletes $n$ consecutive prompt tokens from the beginning of the prompt, and \emph{Back-deletion} deletes $n$ tokens counted backwards from the end of the prompt.
In random-deletion, we ran 100 trials and report the average accuracy.
As in the previous experiments, we used four prompts for AP and report the averaged results.

\begin{table*}[t!]
    \setlength{\tabcolsep}{1.9mm} 
    \centering
    \small
    \begin{tabular}{cccccccccccccc}
    \toprule
    \multirow{2}{*}{\textbf{Task}} & \multirow{2}{*}{\textbf{Method}} & \multirow{2}{*}{\textbf{Metrics}}  & \multicolumn{10}{c}{\textbf{Position of the deleted prompt token}} & \multirow{2}{*}{\textbf{Orig.}} \\
    & & & 1 & 2 & 3 & 4 & 5 & 6 & 7 & 8 & 9 & 10 \\
    \midrule
    \multirow{4}{*}{CB} & 
        \multirow{2}{*}{AP} & Acc & 62.1 & \textbf{61.6} & 63.4 & 59.4 & \uline{65.6} & \uline{65.6} & 62.1 & 63.8 & 62.1 & 62.9 & 68.3\\
        & & RoD & 0.09 & \textbf{0.10} & 0.07 & 0.13 & \uline{0.04} & \uline{0.04} & 0.09 & 0.07 & 0.09 & 0.08 & - \\
        \cmidrule(lr){2-14}
        & \multirow{2}{*}{MP} & Acc & 93.8 & \textbf{93.3} & \uline{96.0} & - & - & - & - & - & - & - & 95.1 \\
        & & RoD & 0.01 & \textbf{0.02} & \uline{-0.01} & - & - & - & - & - & - & - & - \\
        \midrule
    \multirow{4}{*}{MNLI} & 
        \multirow{2}{*}{AP} & Acc & \uline{37.9} & 37.8 & \textbf{36.6} & 37.5 & 37.5 & 37.2 & 37.5 & 37.4 & 37.5 & 37.1 & 37.7 \\
        & & RoD & \uline{-0.01} & 0.00 & \textbf{0.03} & 0.01 & 0.01 & 0.01 & 0.01 & 0.01 & 0.01 & 0.02 & - \\
        \cmidrule(lr){2-14}
        & \multirow{2}{*}{MP} & Acc & 64.5 & \uline{65.4} & \textbf{55.4} & - & - & - & - & - & - & - & 65.5 \\
        & & RoD & 0.02 & \uline{0.00} & \textbf{0.15} & - & - & - & - & - & - & - & - \\
    \bottomrule
    \end{tabular}
    \caption{
    Average accuracy was obtained after deleting a single token at different positions of a given prompt.
    The largest drops in accuracy over the deletion positions are \textbf{bolded} and the smallest drops are \uline{underlined} for each method and dataset.
    Column `Orig.' shows the performance of the original prompt.
    }
    \label{tab:deletion_one}
\end{table*}

\paragraph{Results:}
From \autoref{tab:deletion_one} we see that the \textbf{accuracy of both AP and MP drops even when a single token is deleted} at specific positions.
However, the observed trends differ in CB and MNLI.
For example, AP resulted in higher RoD values in CB compared to MNLI.
This shows that the robustness of AP under single token deletion heavily depends on the dataset.
\autoref{tab:deletion_multi} shows the results for the multiple token deletion setting.
We see that \textbf{the performance of both AP and MP degrades when more tokens are deleted.}
Interestingly, the accuracy drop in CB is very small for MP even when all prompt tokens are deleted (i.e., only the task inputs and \prompt{<MASK>} were used as the input).
This suggests that the performance on CB is less reliant on the prompt tokens in MP.

\subsection{Cross-Dataset Evaluation}
\label{sec:exp-cross-data}
Given that discrete prompt learning methods such as AP learn prompts from a small set of training instances, it is important that the learnt prompts encode generalizable task-specific features and not random artefacts in the training sample used. %
To study the transferability of the learnt discrete prompts from one dataset to another, we conduct a cross-dataset evaluation as described next.

\begin{table}[t!]
    \setlength{\tabcolsep}{0.9mm} %
    \centering
    \small
    \begin{tabular}{ccccccccccccc}
        \toprule
        \multirow{2}{*}{\textbf{Strategy}} & \multirow{2}{*}{\textbf{Method}} & \multirow{2}{*}{\textbf{Metrics}} & \multicolumn{4}{c}{\textbf{\#Deleted Tokens}} & \multirow{2}{*}{\textbf{Orig.}} \\
        & & & 1 & 3 & 5 & 7 \\
        \midrule
        \midrule
        \multicolumn{8}{c}{\textbf{CB}} \\
        \midrule
        \multirow{4}{*}{Random}
            & \multirow{2}{*}{AP}
            & Acc & \uline{56.7} & 56.0 & 55.4 & \textbf{54.8} & 68.3 \\
        & & RoD & \uline{0.17} & 0.18 & 0.19 & \textbf{0.20} & - \\
            \cmidrule(lr){2-8}
            & \multirow{2}{*}{MP}
            & Acc & \textbf{93.3} & \uline{94.6} & - & - & 95.1\\
        & & RoD & \textbf{0.02} & \uline{0.01} & - & - & - \\
        \midrule
        \multirow{4}{*}{Front}
            & \multirow{2}{*}{AP}
            & Acc & \uline{62.1} & \textbf{49.1} & 57.6 & 57.6 & 68.3 \\
        & & RoD & \uline{0.09} & \textbf{0.28} & 0.16 & 0.16 & - \\
            \cmidrule(lr){2-8}
            & \multirow{2}{*}{MP}
            & Acc & \textbf{93.8} & \uline{94.6} & - & - & 95.1\\
        & & RoD & \textbf{0.01} & \uline{0.01} & - & - & - \\
        \midrule
        \multirow{4}{*}{Back}
            & \multirow{2}{*}{AP}
            & Acc & \uline{62.9} & 57.6 & 55.8 & \textbf{51.3} & 68.3 \\
        & & RoD & \uline{0.08} & 0.16 & 0.18 & \textbf{0.25} & - \\
            \cmidrule(lr){2-8}
            & \multirow{2}{*}{MP}
            & Acc & \uline{96.0} & \textbf{94.6} & - & - & 95.1 \\
        & & RoD & \uline{-0.01} & \textbf{0.01} & - & - & - \\
        \midrule
        \midrule
        \multicolumn{8}{c}{\textbf{MNLI}} \\
        \midrule
            \multirow{4}{*}{Random}
            & \multirow{2}{*}{AP}
            & Acc & \textbf{35.8} & \textbf{35.8} & 36.0 & \uline{36.2} & 37.7 \\
        & & RoD & \textbf{0.05} & \textbf{0.05} & 0.05 & \uline{0.04} & - \\
            \cmidrule(lr){2-8}
            & \multirow{2}{*}{MP}
            & Acc & \uline{65.4} & \textbf{52.6} & - & - & 65.5 \\
        & & RoD & \uline{0.0} & \textbf{0.20} & - & - & - \\
        \midrule
        \multirow{4}{*}{Front}
            & \multirow{2}{*}{AP}
            & Acc & \uline{37.9} & 36.5 & 36.2 & \textbf{36.0} & 37.7 \\
        & & RoD & \uline{-0.01} & 0.03 & 0.04 & \textbf{0.05} & - \\
            \cmidrule(lr){2-8}
            & \multirow{2}{*}{MP}
            & Acc & \uline{64.5} & \textbf{52.6} & - & - & 65.5 \\
        & & RoD & \uline{0.02} & \textbf{0.20} & - & - & - \\
        \midrule
        \multirow{4}{*}{Back}
            & \multirow{2}{*}{AP}
            & Acc & \uline{37.1} & 36.7 & 35.7 & \textbf{36.5} & 37.7 \\
        & & RoD & \uline{0.02} & 0.03 & 0.05 & \textbf{0.03} & - \\
            \cmidrule(lr){2-8}
            & \multirow{2}{*}{MP}
            & Acc & \uline{55.4} & \textbf{52.6} & - & - & 65.5 \\
        & & RoD & \uline{0.15} & \textbf{0.20}  & - & - & - \\
        \bottomrule
    \end{tabular}
    \caption{
    Average accuracy was obtained after deleting multiple tokens from a given prompt.
    The largest drops in accuracy over the deleted tokens  are \textbf{bolded} and the smallest drops are \uline{underlined} for each strategy and method.
    }
    \label{tab:deletion_multi}
\end{table}

\paragraph{Experimental Procedure:}
We used one NLI dataset (e.g. CB) to learn the prompts and then use them to make entailment predictions in another NLI dataset (e.g. MNLI).
We then measured the drop in accuracy using RoD for this cross-dataset transferability task with respect to the accuracy of test data from the same dataset.

\paragraph{Results:}
As seen from \autoref{tab:cross-dataset-eval}, \textbf{AP-based prompts do not generalize well across datasets.}
For both AP and MP, RoD is larger in the transfer from CB to MNLI than in the opposite direction.
This implies that MNLI is a better dataset for fine-tuning a PLM for NLI using discrete prompts.

\begin{table}[t]
\centering
\small
\begin{tabular}{lccccc}
\toprule
\multirow{2}{*}{\textbf{Method}}& \multicolumn{2}{c}{\textbf{Test Dataset}} & \multirow{2}{*}{\textbf{RoD}}\\
& \textbf{CB} & \textbf{MNLI} & \\
\midrule
AP trained on CB   & 68.3 & 36.1  & \uline{0.47} \\ 
AP trained on MNLI & 42.9 & 37.7  & \uline{0.12} \\ 
\midrule
MP trained on CB   & 95.1 & 43.4 & \textbf{0.54} \\ 
MP trained on MNLI & 43.8 & 65.5 & \textbf{0.33} \\ 
\bottomrule
\end{tabular}
\caption{
Accuracy and RoD for the cross-dataset evaluation where a method (AP/MP) is trained on one NLI dataset (CB/MNLI) and the learnt prompts are used to make entailment predictions in a different NLI dataset.
}
\label{tab:cross-dataset-eval}
\end{table}

\subsection{Adversarial Perturbations}
\label{sec:exp:input-perturbation}
Introducing carefully designed adversarial perturbations to the test instances such as modifications to sentences that might or might not alter the original target labels have been used as a technique for probing the robustness of models~\cite{DBLP:journals/corr/GoodfellowSS14}. 
Previous studies~\cite{DBLP:journals/corr/SamantaM17,DBLP:conf/aaai/JinJZS20} have shown that pre-trained models can be easily fooled to make incorrect predictions with seemingly innocuous perturbations to the test instances. Therefore, we evaluate discrete prompt-based NLI models for their robustness against adversarially perturbated test instances.

\paragraph{Evaluation Dataset:}
For this purpose, we asked two annotators to manually edit hypothesis sentences in NLI test data considering two types of perturbations:
(1) perturbations that do not change reference labels, and (2) perturbations that change reference labels.
An example is shown in  \autoref{tab:perturbation-data}. 

For the first type of perturbation, we edited a hypothesis sentence such that its relationship with the corresponding premise remains unchanged.
For the second type, we edited a hypothesis sentence such that its relationship (e.g., from \emph{entailment} to \emph{contradiction}) will be reversed.
The premise and hypothesis pairs were sampled from CB (validation set) and MNLI (test set).
Because there are ca. 10,000 test instances in MNLI and it is costly to manually edit sentences, we used 100 randomly-chosen sentence pairs covering MNLI and CB.

\begin{table}[t]
    \centering
    \small
    \begin{tabular}{lp{7em}p{5em}}
        \toprule
        \textbf{} & \textbf{Hypothesis} & \textbf{Label} \\ 
        \midrule
        Original & The Wither's only had daughters. & contradiction \\
        \midrule
        Perturbation & \\
        \quad w/o label changes & The Wither's did not have sons. & contradiction \\
        \quad w/ label changes & The Wither's had a boy. & entailment \\
        \bottomrule
    \end{tabular}
    \caption{
    Examples of our evaluation set consisting of task inputs with perturbations. 
    The premise sentence is `\emph{The Wither's eldest boy, one of the four of the town militia, saluted in the old style with his stick sword.}'
    } 
    \label{tab:perturbation-data}
\end{table}

\paragraph{Experimental Procedure:}
We computed the RoD of average accuracies obtained with original and adversarial test instances.
Specifically, we used the AP prompts in \autoref{tab:pre-train-best} under three settings:
(a) original (without perturbations), 
(b) perturbations without label changes, 
and (c) perturbations with label changes.
Then, we calculate RoD from (a) to (b) and (a) to (c) as shown in \autoref{tab:task-input-perturbation}.

\begin{table}[t!]
\small
\setlength{\tabcolsep}{0.5mm} %
\centering
\begin{tabular}{ccccc}
\toprule
\textbf{Perturbation} & \textbf{Method} & \textbf{Metrics} & \textbf{CB} & \textbf{MNLI} \\
\midrule
\multirow{4}{*}{Original}
    & \multirow{2}{*}{AP} 
      & Acc & 54.5 & 40.5 \\
    & & RoD & - & - \\
    \cmidrule(l){2-5}
    & \multirow{2}{*}{MP} 
      & Acc & 95.5 & 71.0 \\
    & & RoD & - & - \\
\midrule
\multirow{4}{*}{\begin{tabular}{c}Perturbation \\w/o label changes\end{tabular}}
    & \multirow{2}{*}{AP} 
      & Acc & 55.5 & 43.2 \\
    & & RoD & \uline{-0.02} & \uline{-0.07} \\
    \cmidrule(l){2-5}
    & \multirow{2}{*}{MP} 
      & Acc & 93.0 & 66.7 \\
    & & RoD & \textbf{0.03} & \textbf{0.06} \\
\midrule
\multirow{4}{*}{\begin{tabular}{c}Perturbation \\w/ label changes\end{tabular}}
    & \multirow{2}{*}{AP} 
      & Acc & 42.3 & 39.4 \\
    & & RoD & \uline{0.22} & \uline{0.03} \\
    \cmidrule(l){2-5}
    & \multirow{2}{*}{MP} 
      & Acc & 41.8 & 61.2 \\
    & & RoD & \textbf{0.56} & \textbf{0.14} \\
\bottomrule
\end{tabular}
\caption{
Accuracy and RoD in prompts for task inputs that include perturbations. The RoD here is the rate of degradation in the average accuracy from the original without perturbations to perturbations without label changes or perturbations with label changes.
The largest drops in accuracy are \textbf{bolded} and the smallest drops are \uline{underlined} for each perturbation and method.
}
\label{tab:task-input-perturbation}
\end{table}

\paragraph{Results:}
Overall, we see that the RoD of AP is consistently smaller than that of MP in both CB and MNLI under both types of perturbations.
However, it is also clear that the accuracy obtained with AP is much smaller than that with MP.
For the perturbations without label changes, both AP and MP show small RoD values, compared to those with label changes.\footnote{w/o label change modifications slightly increase the average length of a hypothesis and AP seems to better exploit this extra information for inference resulting in a slight improvement in accuracy (negative RoD).}
This shows that both AP and MP are relatively robust against modifications to the hypotheses that do not significantly alter the meaning.
However, when stronger perturbations are introduced that would result in label changes, the accuracy of both AP and MP drops significantly. \footnote{MP is less robust compared to AP, likely as a result of overfitting to strongly perturbed training data during fine-tuning the PLM.}
This is a concern because it shows that \textbf{neither AP nor MP is sufficiently robust to correctly predict the target labels when the hypothesis sentences in test data are adversarially modified.}

\section{Conclusion}
We investigated the robustness of discrete prompts under different perturbations.
We found that although discrete prompts remain relatively robust against token deletion, it is highly sensitive to other types of perturbations such as token shuffling.
For adversarial perturbations to the input, discrete prompts were robust to weak perturbations without label changes, but AP was more robust than MP for perturbations with label changes.
Moreover, they generalize poorly across different datasets annotated for NLI.
We hope our analysis will inspire future work to develop methods that learn both accurate as well as robust discrete prompts.

\section{Limitations}
Possible limitations of this work are:
\begin{itemize}
\item We chose popular discrete prompt methods of AP and MP and did not investigate other methods in this work. Our analysis procedure can still be applied to other discrete prompts such as AvgTrigger~\cite{wallace-etal-2019-universal}.
\item We chose RoBERTa-large following previous studies of HFT~\cite{DBLP:conf/naacl/ScaoR21} and AP~\cite{shin-etal-2020-autoprompt} for reproducible and identical comparisons with them. Other PLMs would lead to different results, but they can also be investigated in the same way as in this work.
\item This work focuses on NLI because it is a fundamental natural language understanding task and still difficult even with PLMs~\cite{GPT3}. Other complex downstream tasks are worth investigating for a deeper understanding of prompt-based approaches in future work.
\item The results and conclusions are from the English datasets and would differ in other languages. However, our methodologies do not depend on English and can be applied to other languages as important future studies. 
\item Since there was a performance gap between MP/HFT and AP, the accuracies by the perturbations could be affected. However, this work does not aim to find the best prompt learning method but to analyze the robustness of discrete prompts for a deeper understanding of them.
\end{itemize}

\section{Ethical Considerations}
Our adversarial dataset came from existing datasets of CB and MNLI.
We visually checked the instances in the data development and found no instances with ethical concerns.

One should also be aware of social biases (e.g. gender stereotypes) in PLM. 
RoBERTa, the PLM we used in our experiments, is known to have gender biases~\cite{DBLP:journals/corr/abs-2105-05541}. 
Since we used it as-is in order to follow the experimental conditions of previous studies using RoBERTa, our current results are possibly influenced by such biases.
However, the consideration of the prompt robustness of this work would not pose or magnify such ethical concerns.

\section{Acknowledgments}
This research was supported by the JSPS KAKENHI (Grants-in-Aid for Scientific Research) JP22H03654.

\bibliography{anthology,custom}
\bibliographystyle{acl_natbib}

\appendix
\section{Supplementary Materials/Appendix}
\label{sec:supplementary}

\autoref{tab:best-ap-prompts-cb} shows the best prompts learnt by AP, which are used in the robustness evaluations.

\begin{table*}[!hbt]
    \setlength{\tabcolsep}{1.5mm} 
    \small
    \centering
    \begin{tabular}{cp{15em}lcc}
        \toprule
        \textbf{Prompt ID} & \textbf{Prompt learnt by AP} & \textbf{Label tokens} & \textbf{Accuracy} \\
        \midrule
        0 & \prompt{\red{p} <MASK> \blue{strikers <MASK>} \blue{<MASK> Ever Want å£« Console} \blue{Encyclopedia Sie ANC} \red{h}} & \begin{tabular}{p{18em}}
                        \textbf{entailment:}  1927, 1897, 1904\\
                        \textbf{contradiction:}  personally, skeptics, squarely\\
                        \textbf{neutral:}  æµ, ä¸Ĭ, ä¹
                        \end{tabular} & 69.64 \\
        \midrule
        1 & \prompt{\red{p} <MASK> \blue{diagnoses undert} \blue{fueling Hist setups prev bound} \blue{advertisers paper records} \red{h}} & \begin{tabular}{p{18em}}
                        \textbf{entailment:}  1930, 1830, 1890 \\
                        \textbf{contradiction:}  contradict, straight, favors \\ 
                        \textbf{neutral:}  à¨, annabin, kb
                        \end{tabular} & 75.00 \\
        \midrule
        2 & \prompt{\red{p} <MASK> \blue{maximize useful} \blue{courts <MASK> malink rooms} \blue{Scrib home interested Service} \red{h}} & \begin{tabular}{p{18em}}
                        \textbf{entailment:}  4000, 1830, THEN \\
                        \textbf{contradiction:}  yet, preferring, Ps \\
                        \textbf{neutral:}  ĭ, Username, ãĥ« \\
                        \end{tabular} & 57.14 \\
        \midrule
        3 & \prompt{\red{p} <MASK> \blue{fever <MASK> <MASK>} \blue{EL <MASK> <MASK> <MASK> ARE ENE} \blue{cue} \red{h}} & \begin{tabular}{p{18em}}
                        \textbf{entailment:}  1890, 1886, 1889 \\
                        \textbf{contradiction:}  yet, endorsing, contradict \\
                        \textbf{neutral:}  ctory, boolean, Boolean \\
                        \end{tabular} & 71.43 \\
        \bottomrule
    \end{tabular}
    \caption{
    Four prompts learnt by AP in CB.
    \red{Red} represents the task inputs, \red{\prompt{h}} represents the hypothesis, \red{\prompt{p}} represents the premise, \blue{blue} represents the prompt tokens (trigger tokens).
    \blue{\prompt{<MASK>}} tokens in the trigger tokens of some prompts are those used to initialize trigger tokens.
    }
    \label{tab:best-ap-prompts-cb}
\end{table*}

\end{document}